\documentclass[letterpaper, 10 pt, conference]{ieeeconf}  % Comment this line out if you need a4paper

\IEEEoverridecommandlockouts                              % This command is only needed if 
\usepackage{graphics} % for pdf, bitmapped graphics files
\usepackage{epsfig} % for postscript graphics files
\usepackage{mathptmx} % assumes new font selection scheme installed
\usepackage{times} % assumes new font selection scheme installed
\usepackage{amsmath} % assumes amsmath package installed
\usepackage{amssymb}  % assumes amsmath package installed
\usepackage{tabularx}
\usepackage{subcaption}
\usepackage[hidelinks]{hyperref}
\usepackage[dvipsnames]{xcolor}

\title{\LARGE \bf
VT-MUSE: Multimodal Unified Sequential Visuotactile Representation Learning for Manipulation
}

\author{VT-MUSE Team \\ {\small \texttt{acondaway@sjtu.edu.cn} and \texttt{qiaochujoshuayang@sjtu.edu.cn}}% <-this % stops a space
\thanks{This paper is supported by Xense Robotics, Shanghai, China. And Fangyuan Shi from BUPT is currently undertaking the internship.}
\thanks{$^*$ means equal contribution as first author, and $^{\dagger}$ means equal contribution as corresponding author.}
}

\begin{document}

\maketitle
\thispagestyle{empty}
\pagestyle{empty}

%%%%%%%%%%%%%%%%%%%%%%%%%%%%%%%%%%%%%%%%%%%%%%%%%%%%%%%%%%%%%%%%%%%%%%%%%%%%%%%%
\begin{abstract}

Contact-rich robotic manipulation requires interpreting local tactile information in the context of global scene geometry. However, utilizing saparately visual and tactile information makes it difficult to offer good representation for robot policy. We propose VT-MUSE, a two-stage framework for learning task-conditioned visuotactile representations before downstream policy training to leverage the complementary modality. The first stage adapts modality-specific encoders through cross-modal and temporal alignment. The second stage learns a conditional Gaussian latent representation from partially masked visual histories and complete tactile histories. With the supervision of masked visual observation reconstruction and tactile depth flow prediction, the representation learns to model both scene context and contact changes. We then integrate the pretrained representation encoder to a lightweight Transformer-based policy through gated cross-attention. Across four simulation benchmark tasks, VT-MUSE achieves a mean success rate of 55.25\%, which improves by 17.75 percentage points. Across four real world tasks, it achieves success rate at 95\%, which achieves 63.75\% improvement over pure tactile injection method. Further result and videos can refer to \href{https://anonymous.4open.science/w/VT-MUSE/}{VT-MUSE project page}.

\end{abstract}

%%%%%%%%%%%%%%%%%%%%%%%%%%%%%%%%%%%%%%%%%%%%%%%%%%%%%%%%%%%%%%%%%%%%%%%%%%%%%%%%
\section{INTRODUCTION}

Contact-rich robotic manipulation is inherently multimodal and partially
observable. Vision provides scene-level geometry, object configuration, and
long-range context, whereas touch directly captures local interaction states
after contact. The two modalities are therefore complementary: tactile feedback can reveal contact events that are visually subtle and can preserve task-relevant information when the manipulated object or end effector becomes occluded \cite{calandra2018more,suresh2024neuralfeels,he2026fawam}. This complementarity is especially important in insertion, reorientation, and compliant manipulation, where small changes in contact geometry may determine whether an action succeeds. Robots must connect these spatial scales, interpreting local contact feedback in relation to the object's configuration, surrounding geometry, and manipulation objective.

\begin{figure}[htbp]
    \centering
    \includegraphics[width=0.9\linewidth]{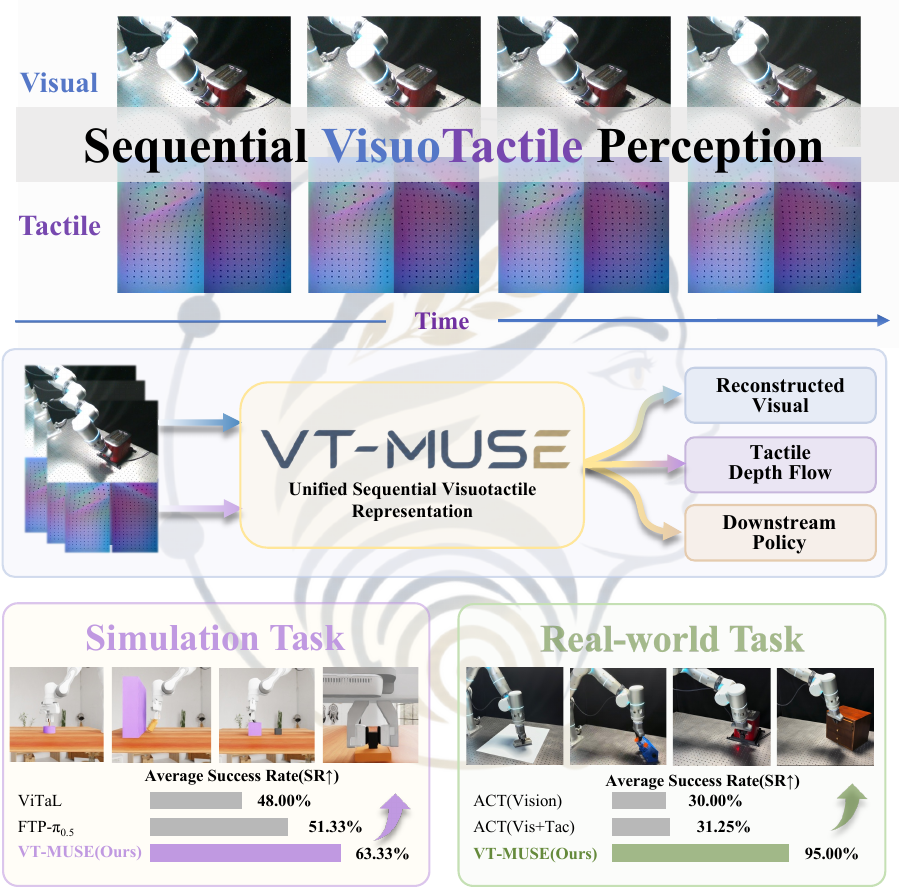}
    \caption{VT-MUSE learns sequential features from complementary visual and tactile histories for contact-rich manipulation. Simulation summaries use LB, PoK, and IHo; physical-robot summaries use four tasks. Full comparisons appear in Tables~\ref{tab:main_results} and~\ref{tab:real_results}.}
    \label{fig:enter-label}
\end{figure}

Recent representation-learning methods have substantially improved the use of
tactile information in manipulation. Self-supervised tactile encoders learn
transferable features across objects, sensors, and tasks
\cite{higuera2024sparsh,zhao2024transferable}, while cross-modal attention and
geometrically aligned representations enable more structured fusion of vision
and touch \cite{chen2022visuo,yuan2024robot}. Temporal information has also been
explored through short force histories, future-observation supervision, tactile
forecasting, and action-conditioned world models
\cite{lee2019making,liu2025vitamin,heng2025vitacformer,zheng2026omnivta}. These advances provide complementary tools for describing sensory content, relating modalities, and modeling temporal change, motivating their integration into a unified perceptual history. Despite this progress, learning what a visuotactile history should preserve for control remains a key challenge. Representation relating local tactile changes to visual context across time while preserving correspondence between the two sensory streams is a must. Longer observation windows leave open which aspects of the interaction the encoder should retain. The learning objective must therefore specify how historical scene information and evolving local contact contribute to the representation. The problem is thus how to retain complementary scene and contact evidence in a compact history representation before downstream action learning.

We address this problem with \textbf{VT-MUSE}, which learns a unified sequential perceptual memory from pooled visuotactile observations before fitting a downstream policy. The \textbf{VT-MUSE Encoder} is trained in two stages with complementary roles. Stage I adapts visual and tactile encoders using cross-modal alignment, temporal consistency, and masked-view consistency to associate observations across modalities and time. Stage II trains a task-conditioned Gaussian latent model on incomplete visual histories and complete tactile histories. Recent tactile tokens query the temporal memory, connecting the readout to ongoing contact while retaining access to earlier scene observations. Reconstructing masked recent visual observations encourages preservation of scene content, while supervising tactile depth changes emphasizes local contact evolution. Masking the visual tail further encourages these predictions to combine tactile evidence with earlier visual context. 

The \textbf{VT-MUSE Policy} combines this memory with current observations through gated cross-attention. By combining visual scene context with evolving tactile evidence, \textbf{VT-MUSE} learns a unified history representation for contact-rich manipulation. Experiments across simulated and physical-robot tasks demonstrate improved manipulation performance over the evaluated baselines. Further ablations show that visual reconstruction and tactile depth flow can complement each other during representation learning, as well as the temporal contrastive supervision can benefit the policy learning. We also find the potential of scale-up of our method to more tasks. To summarize, our contributions are as follows:
\begin{itemize}
    \item A two-stage framework for learning task-conditioned visuotactile history features through cross-modal alignment and multimodal sensory supervision before action learning.
    \item A policy architecture that combines current observations with frozen history features accessed through gated cross-attention.
    \item An evaluation on four simulated and four physical-robot manipulation tasks, with ablations of representation objectives and observation-window length.
\end{itemize}

\section{Related Work}

\subsection{Benchmarks for Visuotactile Manipulation}

Existing resources cover complementary aspects of visuotactile learning.
TacBench evaluates tactile encoders on six touch-centric perception and manipulation tasks \cite{higuera2024sparsh}. At the policy level,
ManiSkill-ViTac defines different challenge tracks for tactile manipulation, while ManiFeel enables controlled comparisons of sensing modalities, tactile representations, and policy architectures \cite{li2024maniskill,luu2025manifeel}. UniVTAC further
combines scalable simulation, supervised representation learning, and an
eight-task manipulation benchmark \cite{chen2026univtac}. Complementary benchmarks study tactile-sensor selection, gentle force-aware manipulation, and whole-hand tactile-guided dexterity \cite{zorin2026taco,wu2026tabero,ni2026tactidex, team2026n_0}.

\subsection{Visuotactile Representation and Policy Learning}

Tactile representation learning has progressed from masked reconstruction and robotic-play pretraining to large-scale self-supervised learning \cite{cao2023learn,higuera2024sparsh}. Cross-sensor learning studies the reuse of tactile features across sensing devices \cite{xu2025unit,zhao2024transferable,tu2026unitac}. Visuotactile methods combine scene and contact information using predictive, contrastive, or attention-based objectives \cite{calandra2018more,lee2019making,chen2022visuo,yuan2024robot,suresh2024neuralfeels}; contrastive pretraining explicitly aligns visual and tactile features \cite{george2025vital,wu2025freetacman}. Complementary policy designs separate global visual reasoning from local visuotactile control \cite{zhao2025touch}.

Temporal prediction is also an established component of visuotactile learning. ViTacFormer combines cross-modal attention with tactile forecasting for action generation \cite{heng2025vitacformer}, while other predictive frameworks model visual, tactile, or action trajectories \cite{zheng2026omnivta,lou2026dream,wu2026tactile,zang2026tacforesight}. Tactile-aware vision--language--action models incorporate tactile representations and prediction objectives into broader policy architectures \cite{cheng2026omnivtla,zhang2026unitacvla,yuan2026ftp,liu2026taco}. VT-MUSE builds on cross-modal fusion and sensory prediction with a focus on representation learning before action prediction: a task-conditioned history encoder is pretrained through sensory objectives, then frozen and accessed by separately trained downstream policies.

\section{Method}

\subsection{Overview}

\begin{figure*}[t]
    \centering
    \includegraphics[width=\linewidth]{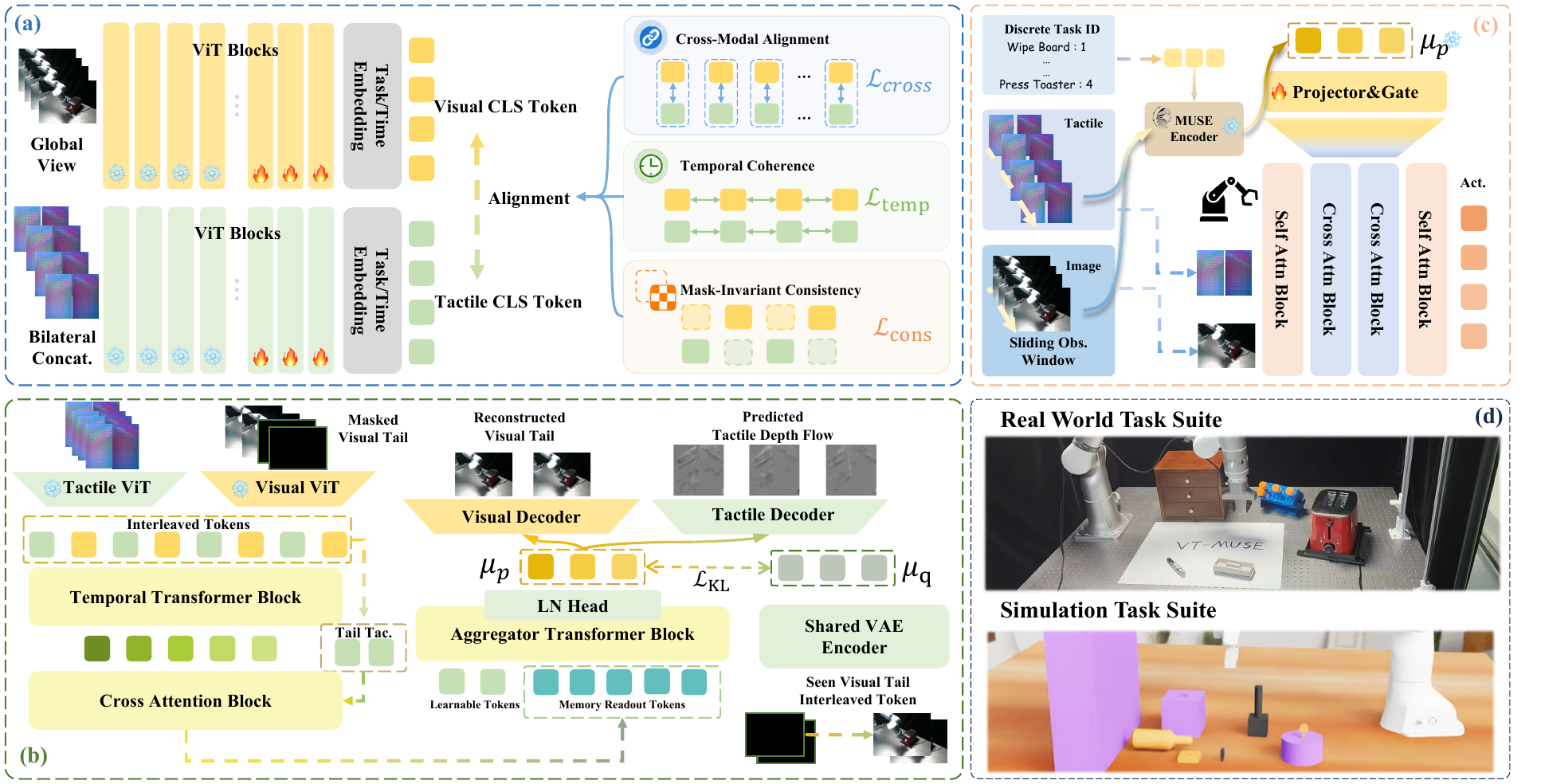}
\caption{\textbf{Overview of VT-MUSE.}
    \color{MidnightBlue}(a) \color{black}Stage I adapts pretrained visual and tactile ViT encoders through cross-modal alignment, temporal consistency, and mask consistency.
    \color{OliveGreen}(b) \color{black}Stage II freezes the modality encoders and processes incomplete visual histories with complete tactile histories. Recent tactile tokens query the temporal memory. Auxiliary decoders reconstruct masked visual observations and tactile depth changes; the training-only posterior additionally receives the unmasked visual-tail tokens.
    \color{YellowOrange}(c) \color{black}The frozen prior features are projected into the policy hidden space and supplied to intermediate layers through gated cross-attention.
    \color{Gray}(d) \color{black}Evaluation uses physical-robot tasks and the UniVTAC simulation benchmark.}

    \label{fig:pipeline}
\end{figure*}

As shown in Fig.~\ref{fig:pipeline}, VT-MUSE combines a pretrained sequential
perceptual encoder with a Transformer-based action-chunking policy.
Representation learning proceeds in two stages. Stage I aligns visual and
tactile observations while enforcing temporal and masking consistency. Stage II
learns a conditional latent representation by reconstructing masked visual
observations and predicting changes in tactile geometry. The resulting
representation is frozen and provided to the VT-MUSE Policy as a separate
memory through gated cross-attention. This separation allows the VT-MUSE Encoder
to learn from pooled multi-task data, while each downstream policy can be
optimized using a smaller task-specific demonstration set.

\subsection{Problem Formulation}

We consider a multi-task robotic manipulation dataset consisting of trajectories
\(\tau=\{(o_t,a_t)\}_{t=1}^{T}\), where
\(o_t = \left(I_t,T_t,q_t,c\right)\). Here, \(I_t\in\mathbb{R}^{3\times H_I\times W_I}\) denotes the external RGB
observation, and \(T_t=(T_t^{L},T_t^{R})\) contains the RGB observations from
the left and right optical tactile sensors. The variable \(q_t\) denotes the
robot proprioceptive state, \(a_t\) is the corresponding robot action, and
\(c\) is a discrete task identity used during multi-task representation
learning. For each target time step \(t\), we construct a temporally subsampled
observation window of length \(L\) with sampling stride \(s\):
\(
\mathbf{I}_t =
\left(
I_{t-(L-1)s},\ldots,I_{t-s},I_t
\right),
\mathbf{T}_t =
\left(
T_{t-(L-1)s},\ldots,T_{t-s},T_t
\right)
\). The VT-MUSE Encoder is designed as a perceptual state encoder. During representation learning, selected visual observations near the end of the window are masked, producing a partially observed visual sequence \(\bar{\mathbf{I}}_t\). Given this incomplete visual history, the synchronized tactile history, and the task identity, the encoder infers a temporally grounded visuotactile representation:
\begin{equation}
f_t^{\mathrm{VT}}
=
E_{\mathrm{VT}}
\left(
\bar{\mathbf{I}}_t,
\mathbf{T}_t,
c
\right).
\end{equation}

The objective of the VT-MUSE Encoder is to provide the downstream policy with a temporally consistent and visuotactile-grounded representation. This
representation is supplied to the VT-MUSE Policy together with the current
visual, tactile, and proprioceptive observations. The policy predicts an action chunk with horizon \(K_A\):
\begin{equation}
\hat{\mathbf{A}}_t
=
\pi
\left(
q_t,
I_t,
T_t,
f_t^{\mathrm{VT}}
\right)
=
\left[
\hat{a}_t,
\hat{a}_{t+1},
\ldots,
\hat{a}_{t+K_A-1}
\right].
\end{equation}

Thus, VT-MUSE maintains two complementary sources of information: a historical
visuotactile window summarized by the VT-MUSE Encoder and the current sensory
observations directly available to the VT-MUSE Policy. The former provides
temporally aggregated interaction context, while the latter supports immediate
closed-loop action prediction.

\subsection{Model Structure}

\subsubsection{Temporal Visuotactile Encoding}

We use separate visual and tactile Vision Transformer (ViT) encoders initialized from pretrained weights. Each observation contributes one CLS token. The left and right tactile RGB images are horizontally concatenated and resized into a single three-channel observation. Tokens receive modality, temporal-slot, and relative-time embeddings.

For Stage II, a nonempty subset of the final \(K_V\) visual tokens is replaced with learnable mask tokens, while tactile observations remain available. We uniformly samples three patterns: masking the penultimate token, the final token, or both. Here, \(K_V\) is distinct from the action horizon \(K_A\). Masking precedes task conditioning: a learnable task query reads the masked visual and complete tactile sequence, and its context is added to both modalities. Their tokens are interleaved in temporal order and processed into memory \(M\). Task-conditioned readout produces \(r_c\), which enriches the final \(K_V\) tactile queries \(Q\). Cross-attention yields \(C=\operatorname{CrossAttn}(Q,M,M)\); the context \(C\), queries \(Q\), and \(r_c\) are aggregated into a latent representation.

Recent tactile observations provide local evidence about the current contact state. Using their tokens as queries lets the readout combine this recent evidence with the visual and tactile context stored across the window. The resulting representation brings together scene-level observations and contact history before it is exposed to the action policy.

VT-MUSE uses a conditional prior for downstream feature extraction and a privileged posterior during representation training. Let \(\mathcal{C}_t=(\bar{\mathbf{I}}_t,\mathbf{T}_t,c)\) denote the prior's sensory and task inputs, and let \(Y_t\) contain the ground-truth visual tokens for all \(K_V\) tail slots. The two branches parameterize diagonal Gaussian latent distributions separately.

Both branches share encoder, latent-aggregation, and Gaussian-head parameters. Only the posterior appends the additional target tokens to the latent aggregator. The prior receives visual-tail content only at unmasked positions and is used for reconstruction and downstream feature extraction.

\subsubsection{Auxiliary Multimodal Reconstruction Heads}

A shared image decoder with frame-specific embeddings reconstructs the masked RGB tail. An independent tactile decoder reconstructs depth changes between adjacent sampled observations. Let \(D_i^b\) denote the tactile depth map at sampled slot \(i\) on side \(b\in\{L,R\}\). The depth flow target is \(\Delta D_i^b=D_i^b-D_{i-1}^b\).
The first sampled slot uses a zero difference. The left and right depth flow maps are horizontally concatenated
and resized to form the reconstruction target \(d\). With \(u=\operatorname{clip}(d/\delta_{\max},-1,1)\), the three-channel target is \(G(d)=((1+u)/2,\max(u,0),\max(-u,0))\). The term \emph{tactile depth flow} in the figures refers to this encoding of observed depth changes. RGB reconstruction supervises scene appearance, while depth flow reconstruction supplies local contact supervision.

\subsubsection{Action Policy}

After representation learning, we retain the conditional-prior encoder and use its mean \(\mu_p\) as the history feature \(f_t^{\mathrm{VT}}\). A two-layer adapter with layer normalization and GELU activation projects this feature into the policy hidden space. The projected representation is supplied as cross-attention memory, alongside the policy's current sensory and proprioceptive observations. Let \(X_{\mathrm{act}}\) denote the policy action tokens and \(U_{\mathrm{VT}}\) the projected history features. The fused tokens are

\begin{equation}
X_{\mathrm{act}}'
=
X_{\mathrm{act}}
+
g\,
\operatorname{CrossAttn}
\left(
X_{\mathrm{act}},
U_{\mathrm{VT}},
U_{\mathrm{VT}}
\right),
\end{equation}
where \(g\) is a learnable scalar gate. The gate parameter is initialized such that the auxiliary representation has only a small influence at the beginning of policy training. The residual form retains the policy action tokens while adding a contribution from the frozen historical features. This allows task-specific action learning to determine how strongly those features enter the policy, without updating the pretrained history encoder.

\begin{figure*}[h]
    \centering
    \includegraphics[width=\linewidth]{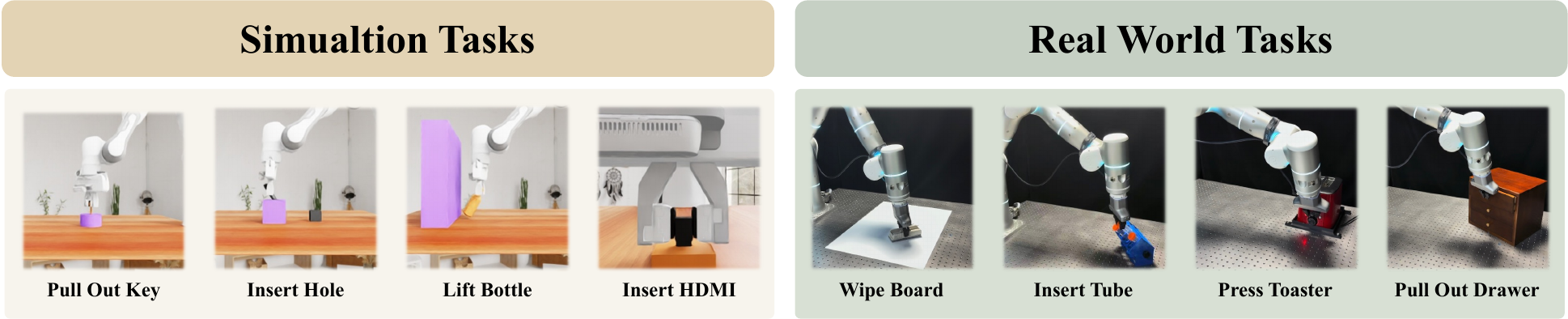}
    \caption{\textbf{Evaluation tasks.} VT-MUSE is evaluated on four simulated UniVTAC tasks and four physical-robot tasks covering grasping, compliant interaction, contact-based exploration, and precision insertion.}
    \label{fig:tasks}
\end{figure*}

\subsection{Training Objective}

\subsubsection{Stage I, Cross-Modal Temporal Alignment}

Stage I adapts the pretrained modality encoders to robot interaction data.
The training objective combines synchronous visual--tactile alignment, temporal-state contrast, and masked-view consistency. Alignment operates on tokens enriched with a shared task-conditioned multimodal context. Synchronized tokens form positive pairs; other batch samples, gathered across devices in distributed training, provide negatives. We use symmetric InfoNCE:
\begin{equation}
\mathcal{L}_{\mathrm{cross}}
=
\frac{1}{L}
\sum_{i=1}^{L}
\frac{1}{2}
\left(
\ell_{v\rightarrow h}^{(i)}
+
\ell_{h\rightarrow v}^{(i)}
\right),
\end{equation}

where \(\ell_{v\rightarrow h}^{(i)}\) and \(\ell_{h\rightarrow v}^{(i)}\) denote the visual-to-tactile and tactile-to-visual contrastive losses, respectively.

For each temporal slot, we average the synchronized visual and tactile tokens and normalize the resulting multimodal state, denoted by \(s_i\).

Temporal coherence is encouraged through contrastive learning between adjacent sampled states:
\begin{equation}
\mathcal{L}_{\mathrm{temp}}
=
\frac{1}{L-1}
\sum_{i=1}^{L-1}
\mathcal{L}_{\mathrm{NCE}}
\left(
s_i,
s_{i+1}
\right).
\end{equation}
This objective encourages temporally neighboring states from the same trajectory to remain close in the representation space.

% \subsubsection{Mask Consistency}

For the same observation window, we independently sample two masks, \(m^{(1)}\) and \(m^{(2)}\), each masking every visual slot independently with set probability.
Let \(s_i^{(1)}\) and \(s_i^{(2)}\) denote the corresponding multimodal representations.
We minimize the cosine consistency loss
\begin{equation}
\mathcal{L}_{\mathrm{cons}}
=
1
-
\frac{1}{BL}
\sum_{b=1}^{B}
\sum_{i=1}^{L}
\cos
\left(
s_{b,i}^{(1)},
s_{b,i}^{(2)}
\right).
\end{equation}
Here, \(B\) is the batch size. This objective encourages consistency across the sampled visual masks. The complete Stage I objective is
\begin{equation}
\mathcal{L}_{\mathrm{Stage1}}
=
\lambda_{\mathrm{cross}}
\mathcal{L}_{\mathrm{cross}}
+
\lambda_{\mathrm{temp}}
\mathcal{L}_{\mathrm{temp}}
+
\lambda_{\mathrm{cons}}
\mathcal{L}_{\mathrm{cons}}.
\end{equation}

During Stage I, we fine-tune the final three Transformer blocks
and the final layer normalization of each modality encoder,
together with the task-conditioning and temporal-embedding modules.
All three loss terms are equally weighted.

\subsubsection{Stage II, Latent Modeling}

Stage II freezes the Stage-I modality encoders and trains the temporal memory, tactile-query readout, shared Gaussian latent model, and auxiliary decoders. Both decoders reconstruct from the deterministic prior mean \(z=\mu_p(\mathcal{C}_t)\); no latent sampling is used. RGB and normalized depth-target mean squared errors, \(\mathcal{L}_{\mathrm{rgb}}\) and \(\mathcal{L}_{\mathrm{depth}}\), are averaged only over visually masked tail slots. The privileged posterior enters through the distributional regularizer.

So the complete objective is
\begin{equation}
\mathcal{L}_{\mathrm{Stage2}}
=\lambda_I\mathcal{L}_{\mathrm{rgb}}
+\lambda_D\mathcal{L}_{\mathrm{depth}}
+\beta\mathcal{L}_{\mathrm{KL}}.
\end{equation}
The KL term is summed over latent dimensions and averaged over the batch, with gradients through both branches. Thus, the objective combines prior-mean reconstruction with posterior-prior regularization. 
% The released script uses \(\lambda_I=\lambda_D=1\), \(\beta=10^{-3}\), and zero perceptual-loss weight.

\subsubsection{Policy Training}

During downstream policy training, the Stage-II conditional-prior encoder remains frozen. We use a four-layer Transformer policy with cross-attention to history features at two intermediate layers. An \(L_1\) loss between the predicted and demonstrated action chunks, \(\hat{\mathbf{A}}_t\) and \(\mathbf{A}_t\), defines \(\mathcal{L}_{\mathrm{act}}\). The policy's action posterior is regularized toward a standard Gaussian through \(\mathcal{L}_{\mathrm{act\mbox{-}KL}}\). This action latent is separate from the sensory latent used for history representation learning. The policy objective is

\begin{equation}
\mathcal{L}_{\mathrm{policy}}
=
\mathcal{L}_{\mathrm{act}}
+
\lambda_{\mathrm{act}}
\mathcal{L}_{\mathrm{act\mbox{-}KL}}.
\end{equation}

% The detailed formal explanation is attached in the Appendix.

\section{Experiments}

We organize the experiments around three questions:

\textbf{Q1: Downstream effectiveness.} How does the complete VT-MUSE pipeline perform relative to the evaluated policy configurations and published reference results?

\textbf{Q2: Multimodal supervision.} How do the visual and tactile reconstruction objectives affect downstream success and auxiliary prediction quality within VT-MUSE?

\textbf{Q3: Temporal configuration.} How do the adjacent-state contrastive objective and observation-window length affect performance?

\subsection{Experimental Setup}

\subsubsection{Tasks and Data}

We evaluate VT-MUSE on four simulated and four physical-robot manipulation
tasks, as illustrated in Fig.~\ref{fig:tasks}. For simulation, we use four tasks
from the UniVTAC platform \cite{chen2026univtac}: Lift Bottle (LB), Pull-out Key
(PoK), Insert Hole (IHo), and Insert HDMI (IHD). We collect 500 trajectories per
task, resulting in 2,000 trajectories for representation learning. For
downstream policy learning, we use the 50 demonstrations per task released with
the benchmark.

The simulation tasks cover complementary contact patterns. In LB, the robot
grasps a horizontally placed bottle and lifts it while keeping its base within
5\,cm of a wall. In PoK, the robot rotates a key until mechanical resistance is
encountered and then withdraws it along a straight path. IHo requires
contact-based exploration of the hole orientation before tube insertion,
whereas IHD requires fine connector alignment and insertion under rotational
uncertainty.

For the physical-robot experiments, we use a Flexiv Rizon 4s arm equipped with
an XenseGripper. Demonstrations are collected through force-feedback
teleoperation using the Flexiv RDK and TDK toolkits. We consider four tasks:
Insert Tube (IT), Wipe Board (WB), Pull-out Drawer (PoD), and Press Toaster
(PT). IT requires alignment and insertion under object-pose and grasp
variations, including alignment with two holes: one visible to the external camera and one requiring contact feedback. WB requires maintaining surface contact while removing a marker trace from a magnetic board attached to the table; the trace location varies within a designated region. PoD requires grasping the drawer handle and pulling it to the target state, while PT requires reliably actuating the toaster control through
contact. We collect 50 demonstrations per task, resulting in 200 real-world
trajectories. The pooled sensory sequences from these demonstrations train the representation model, while each task-specific policy uses that task's demonstrations and action labels. Thus, representation learning and policy learning use observations from the same task set.

\subsubsection{Training and Evaluation Protocol}

For the simulated experiments, the two representation-learning stages require
approximately 25\,h in total on eight NVIDIA A800 GPUs. The resulting VT-MUSE
Encoder is frozen during downstream policy learning. Each task-specific
VT-MUSE Policy is trained for approximately 30\,min on one NVIDIA A800 GPU
using 50 demonstrations. For locally trained methods, we evaluate each policy
on different 100 seeds per task and report rollout success rate
(SR). Our simulation evaluations run on an NVIDIA RTX 4090 GPU.

For the physical-robot experiments, each method is evaluated for 20 trials per
task under the same task-specific initialization and success criteria. All
policies are deployed on a single NVIDIA RTX 5060 GPU with 8\,GB of memory.

\textit{Released configuration.} The public scripts specify \(L=5\), stride \(s=5\) recorded steps, a 512-dimensional latent, and four layers each for temporal memory and latent aggregation. Stage I and II run for 30 and 20 epochs, respectively, using AdamW with learning rate \(10^{-4}\), weight decay \(0.01\), and cosine decay. Training and validation use separately supplied trajectory directories.

\subsubsection{Baselines}

Our simulation comparisons include \textbf{ACT} \cite{zhao2023act}, a vision-only action-chunking Transformer; \textbf{ACT+UniVTAC}, which uses the representation released with the benchmark; and \textbf{VITaL Pretraining} \cite{george2025vital}, which contrastively pretrains visual and tactile encoders. The VITaL scores are taken from the UniVTAC benchmark evaluation \cite{chen2026univtac}.

We also include published \textbf{FTP-$\pi_{0.5}$} and \textbf{FTP-1} results \cite{yuan2026ftp,intelligence2025pi_}. FTP-$\pi_{0.5}$ uses the FTP architecture initialized in part from $\pi_{0.5}$ without FTP-1 tactile pretraining; FTP-1 additionally uses large-scale tactile pretraining. These results cover LB, PoK, and IHo among our four tasks.

For the physical-robot experiments, we train two ACT baselines using the same
50 demonstrations per task as VT-MUSE. \textbf{ACT (Vision)} receives only the
current external RGB observation, whereas \textbf{ACT (Visuotactile)} directly
receives the current RGB and tactile observations. Neither physical-robot
baseline uses the UniVTAC Encoder or another pretrained representation.

\subsection{Downstream Manipulation Performance (Q1)}

\subsubsection{Simulation Results}

Table~\ref{tab:main_results} reports success rates on UniVTAC. VT-MUSE achieves a four-task average SR of \(55.25\%\), compared with \(39.00\%\) for ACT+UniVTAC, an improvement of \(16.25\) percentage points. On the three tasks shared with the FTP results, VT-MUSE obtains \(63.33\%\), compared with \(51.33\%\) for FTP-$\pi_{0.5}$ and \(69.67\%\) for FTP-1. The FTP comparisons involve different pretraining data and resources.

The gains vary across tasks. Relative to ACT+UniVTAC, VT-MUSE improves IHo from \(23\%\) to \(68\%\) and IHD from \(17\%\) to \(31\%\), while PoK decreases from \(45\%\) to \(38\%\). These results support the effectiveness of the complete pipeline on the evaluated task set, with task-dependent gains that are largest on the two insertion tasks.

IHo and IHD combine geometric alignment with contact-guided insertion, whereas LB involves maintaining a grasp during lifting. These tasks expose different requirements for scene-level and local contact information. VT-MUSE improves LB by 13 percentage points over ACT+UniVTAC in addition to its insertion gains. PoK remains a challenging counterexample, motivating attention to task-level performance alongside the aggregate score.

\begin{table}[htbp]
\centering
\caption{Success rate (SR, \%) on UniVTAC. ``Avg.'' uses all four tasks; ``Avg. (3)'' uses LB, PoK, and IHo. $^\dagger$ marks published results: VITaL from UniVTAC \cite{chen2026univtac}, and FTP variants from \cite{yuan2026ftp}. Bold denotes the highest entry per column across these training regimes.}

\label{tab:main_results}

\renewcommand{\arraystretch}{1.25}
\setlength{\tabcolsep}{3pt}
\small

\begin{tabular*}{\columnwidth}{@{\extracolsep{\fill}}lcccccc@{}}
\hline
Method & LB & PoK & IHo & IHD & Avg. & Avg. (3) \\
\hline
ACT (Pure Vision) & 42 & 28 & 19 & 15 & 26 & 30.00 \\
ACT+UniVTAC & 71 & 45 & 23 & 17 & 39 & 46.33 \\
VITaL$^\dagger$ & 72 & 47 & 25 & 6 & 37.5 & 48.00 \\
FTP-$\pi_{0.5}$$^\dagger$ & 77 & 30 & 47 & - & - & 51.33 \\
FTP-1$^\dagger$ & \textbf{97} & \textbf{48} & 64 & - & - & \textbf{69.67} \\
\hline
VT-MUSE (Ours) & 84 & 38 & \textbf{68} &
\textbf{31} & \textbf{55.25} & 63.33 \\
\hline
\end{tabular*}
\end{table}

\begin{table}[htbp]
\centering
\caption{Physical-robot results. Entries report successful trials out of 20,
and ``Avg.'' is the aggregate success rate over all 80 trials. Higher is
better.}
\label{tab:real_results}

\renewcommand{\arraystretch}{1.25}
\setlength{\tabcolsep}{3pt}
\small

\begin{tabular*}{\columnwidth}{@{\extracolsep{\fill}}lccccc@{}}
\hline
Method & IT & WB & PoD & PT & Avg.  \\
\hline
ACT (Pure Vision) & 1/20 & 3/20 & 13/20 & 4/20 & 26.25 \\
ACT (Vision and Tactile) & 1/20 & 5/20 & 15/20 & 4/20 & 31.25 \\
\hline
VT-MUSE (Ours) & 19/20 & 19/20 & 20/20 & 18/20 & \textbf{95.00} \\
\hline
\end{tabular*}
\end{table}

\begin{figure*}[htbp]
    \centering
    \includegraphics[width=0.8\linewidth]{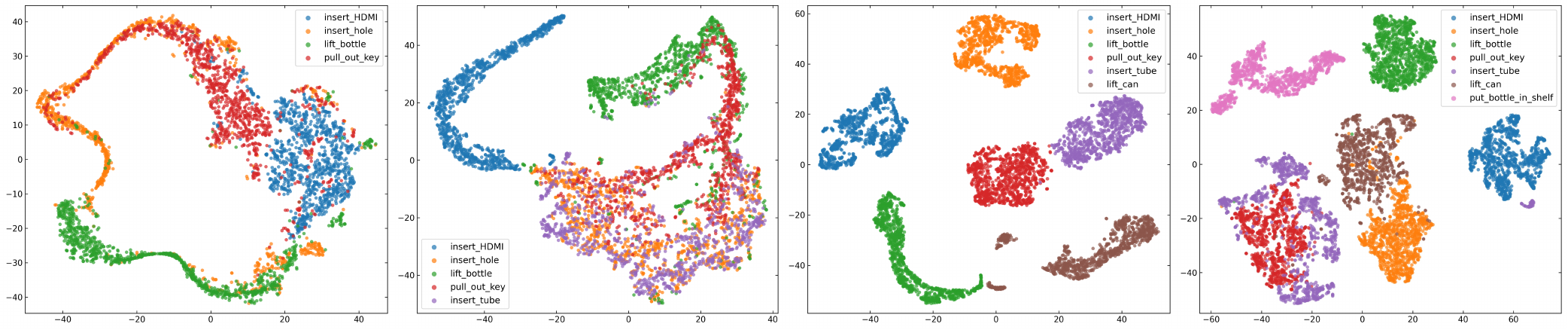}
    \caption{\textbf{Qualitative analysis of multi-task pretraining.}
t-SNE projections of VT-MUSE representations as the pretraining set expands
from four to seven tasks. Each point denotes a temporal observation window, and
colors indicate task identities.}
    \label{fig:scaling}
\end{figure*}

\subsubsection{Physical-Robot Results}

Table~\ref{tab:real_results} reports the physical-robot results. VT-MUSE succeeds in 76 of 80 trials, yielding an aggregate SR of \(95.00\%\), compared with \(26.25\%\) for ACT (Vision) and \(31.25\%\) for ACT (Visuotactile). It exceeds the latter by \(63.75\) percentage points and achieves at least \(18/20\) successes on every task. These results support the effectiveness of the complete representation-to-policy pipeline under the evaluated conditions. The comparison jointly reflects differences in representation pretraining, observation history, and policy integration. With 20 trials per task, the results provide physical-robot validation within this evaluation protocol.

The largest absolute gain over ACT (Visuotactile) occurs on IT, which increases from 1 to 19 successful trials. WB and PT each obtain 14 additional successes, while PoD improves from 15 to 20 successes. The improvements therefore span insertion, sustained surface contact, pulling, and control actuation. This task-level view complements the aggregate success rate by showing how the complete pipeline performs across different physical interactions.

\subsection{Multimodal Objective Ablations (Q2)}

We investigate how the modality-specific Stage-II objectives affect downstream manipulation. All variants retain both visual and tactile inputs and use the same policy architecture, policy demonstrations, and evaluation protocol. The ablation names refer to removed training objectives rather than removed sensory modalities.

As shown in Table~\ref{tab:ablation_results}, removing tactile depth flow supervision reduces average SR from \(55.25\%\) to \(38.75\%\), a drop of \(16.50\) percentage points. Removing visual reconstruction reduces SR to \(37.00\%\), a drop of \(18.25\) points. Both comparisons support the utility of these objectives within VT-MUSE.

The Stage-I-only representation reaches \(27.25\%\). Because this variant omits Stage II as a whole, it measures the contribution of the complete second training stage. The full model performs better than either single-objective variant, consistent with using visual appearance and local contact changes as complementary sources of sensory supervision.

At the task level, the full model exceeds both single-objective variants on all four tasks. IHo shows the largest differences: 31 points relative to removing tactile supervision and 38 points relative to removing visual reconstruction. This pattern is consistent with the design motivation of retaining both scene appearance and contact changes when constructing the history representation for insertion.

\begin{table}[htbp]
\centering
\caption{Auxiliary prediction quality on UniVTAC. Arrows indicate whether
higher or lower values are better. Bold denotes the best result in each
column.}
\label{tab:ablation_results_vq}

\renewcommand{\arraystretch}{1.25}
\setlength{\tabcolsep}{3pt}
\small

\begin{tabular*}{\columnwidth}{@{\extracolsep{\fill}}lcccc@{}}
\hline
Method & PSNR$\uparrow$ & SSIM$\uparrow$ & Vis. MSE$\downarrow$ & Dep. MSE$\downarrow$\\
\hline
w/o Vis  & 9.65 & 0.019 & 0.1084 & 0.007028 \\
w/o Tac & \textbf{33.25} & \textbf{0.983} & \textbf{0.0007932} & 0.1792 \\
w/o TempLoss & 31.47 & 0.981 & 0.000859 & 0.009459 \\
\hline
Ours & 32.53 & 0.981 & 0.0008665 & \textbf{0.006222} \\
\hline
\end{tabular*}
\end{table}

\begin{figure*}[htbp]
    \centering
    \includegraphics[width=0.8\linewidth]{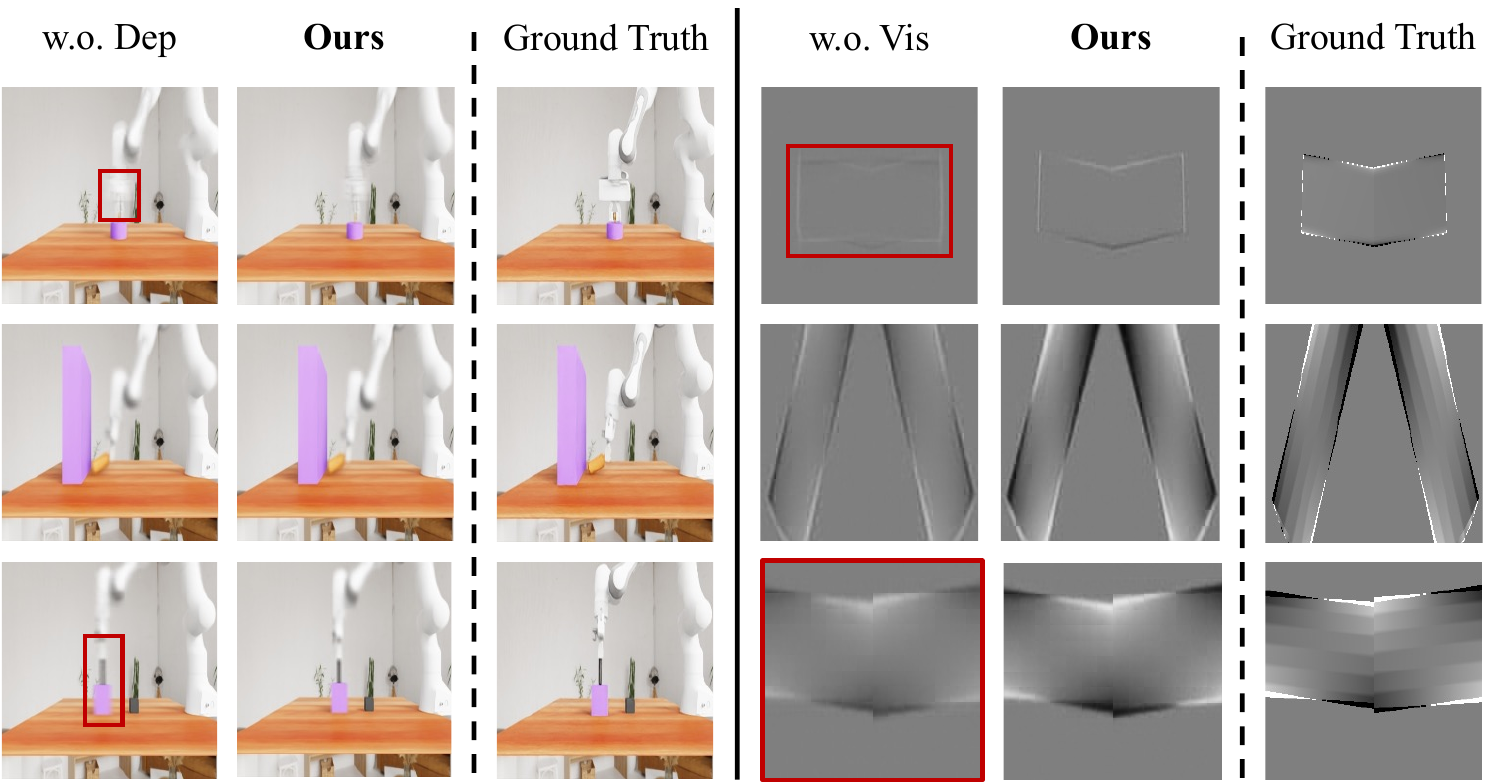}
    \caption{\textbf{Qualitative auxiliary predictions.}
    Visual reconstruction and tactile depth flow reconstruction under different
    representation-learning objectives. Ground-truth observations are shown
    for reference.}
    \label{fig:recon}
\end{figure*}

\subsubsection{Auxiliary Prediction Diagnostics}

Table~\ref{tab:ablation_results_vq} reports auxiliary prediction quality. The variant without tactile depth flow supervision achieves the best visual reconstruction scores, whereas the full model has the lowest tactile depth flow error. The former nevertheless achieves only \(38.75\%\) downstream SR, compared with \(55.25\%\) for VT-MUSE. Better visual reconstruction therefore does not necessarily correspond to better manipulation performance across these variants. We use reconstruction quality as an auxiliary diagnostic and rollout success as the direct measure of policy performance.

Figure~\ref{fig:recon} illustrates selected auxiliary outputs, showing visual object boundaries and tactile depth flow patterns under the different objectives. These examples complement the aggregate reconstruction metrics; the downstream effects of the objectives are evaluated separately in Table~\ref{tab:ablation_results}.
The evaluator uses prior-mean reconstructions and scores visually masked tail frames from the validation loader. PSNR and SSIM are computed per frame and then averaged.

\begin{table}[htbp]
\centering
\caption{Ablation of VT-MUSE representation-learning objectives on UniVTAC.
``w/o Vis'' removes visual reconstruction, ``w/o Tac'' removes tactile depth flow reconstruction, and ``Stage I only'' omits Stage II. ``w/o TempLoss'' removes only the adjacent-state temporal contrastive loss while retaining the temporal Transformer and all other objectives. Success rates are reported in percent.}
\label{tab:ablation_results}

\renewcommand{\arraystretch}{1.25}
\setlength{\tabcolsep}{3pt}
\small

\begin{tabular*}{\columnwidth}{@{\extracolsep{\fill}}lccccc@{}}
\hline
Method & LB & PoK & IHo & IHD & Avg. \\
\hline
w/o Tac  & 75 & 22 & 37 & 21 & 38.75 \\
w/o Vis & 80 & 20 & 30 & 18 & 37 \\
Stage I only & 55 & 15 & 25 & 14 & 27.25 \\
w/o TempLoss & 64 & 15 & 63 & 27 & 42.25 \\
\hline
Ours & \textbf{84} & \textbf{38} & \textbf{68} &
\textbf{31} & \textbf{55.25} \\
\hline
\end{tabular*}
\end{table}

\subsection{Temporal Objective and Window Length (Q3)}

We evaluate the adjacent-state temporal contrastive objective using the \emph{w/o TempLoss} variant. Only this Stage-I loss is removed; the temporal Transformer, visual reconstruction, tactile depth flow supervision, and other training components are retained.

Removing the loss reduces average SR from \(55.25\%\) to \(42.25\%\), a drop of \(13.00\) percentage points. The decrease is task-dependent: 20 points on LB, 23 on PoK, 5 on IHo, and 4 on IHD. With the temporal architecture unchanged, this comparison supports the usefulness of adjacent-state contrastive supervision within the evaluated configuration.

\subsubsection{Observation-Window Length}

Table~\ref{tab:window_results} reports sensitivity to observation-window length at a fixed sampling stride. Among the tested settings, \(L=5\) achieves the highest average SR of \(55.25\%\), exceeding \(L=4\) by \(15.25\) percentage points and \(L=6\) by \(14.00\) points. Performance is non-monotonic over the tested range, indicating sensitivity to the selected history length. The other experiments use \(L=5\).

The five-observation setting also achieves the highest score on every individual task, so its average advantage is not driven by a single task. Extending the window to six observations reduces all four scores. At a fixed stride, window length changes both the number of observations and their temporal coverage; the results identify a useful operating point for the evaluated configuration.

\begin{table}[htbp]
\centering
\caption{Effect of temporal observation-window length on UniVTAC success rate (SR, \%). The sampling stride is held fixed across all settings.}
\label{tab:window_results}

\renewcommand{\arraystretch}{1.25}
\setlength{\tabcolsep}{3pt}
\small

\begin{tabular*}{\columnwidth}{@{\extracolsep{\fill}}lccccc@{}}
\hline
Window Length & LB & PoK & IHo & IHD & Avg. \\
\hline
3 & 60 & 16 & 56 & 14 & 36.5 \\
4 & 60 & 23 & 57 & 20 & 40 \\
\hline
5 (Ours) & 84 & 38 & 68 & 31 & \textbf{55.25} \\
6 & 68 & 18 & 61 & 18 & 41.25 \\
\hline
\end{tabular*}
\end{table}

\subsection{Qualitative Analysis of Multi-Task Pretraining}

We add three UniVTAC tasks---Lift Can, Put Bottle in Shelf, and Insert Tube---and visualize representations learned from four to seven tasks in Fig.~\ref{fig:scaling}. Each point corresponds to an observation window, colored by task identity. Because task identity is an encoder input and t-SNE is a qualitative projection, these plots are used to inspect task-conditioned feature organization. They do not quantify transfer to unseen tasks or changes in downstream performance as the pretraining set expands.

\section{CONCLUSIONS}

VT-MUSE learns task-conditioned visuotactile history features through cross-modal alignment and sensory reconstruction. A task-specific action-chunking policy accesses the frozen encoder through gated cross-attention. Experiments on four simulated and four physical-robot tasks support the complete pipeline, while ablations show the utility of the auxiliary objectives and temporal contrastive supervision, and sensitivity to observation-window length.

\textbf{Limitations and future work.} Evaluation focuses on tasks represented in pretraining and uses a fixed sampling stride and history length. The system comparisons jointly reflect pretraining, historical inputs, and memory integration. Generalization to unseen tasks and sensors, robustness to controlled visual corruption, and variability across independent training runs remain to be assessed. Future work will study adaptive history selection and broader evaluation.

\section{ACKNOWLEDGEMENT}

\begin{itemize}
    \item Project Leaders: Congsheng Xu$^{1}$, Qiaochu Yang$^{1}$
    \item Project Core Members: Fangyuan Shi$^{2}$, Yifan Han$^{1}$, Baijun Chen$^{1}$, Yiming Wang$^{1,2}$, Haonan Zhao$^{1,2}$
    \item Mentors: Yao Mu$^{1}$, Zhe Liu$^{1}$,
    \item Corresponding Author: Daolin Ma$^{1,2}$, Xiaokang Yang$^{1}$, Hesheng Wang$^{1}$,
\end{itemize}

$^{1}$represents Shanghai Jiao Tong University, $^{2}$represents Xense Robotics. Congsheng Xu was primarily responsible for forming model structure and code, robotic infrastructure, and manuscript writing. Qiaochu Yang was primarily responsible for data collection and executing model training and experiments.

\addtolength{\textheight}{-12cm}   % This command serves to balance the column lengths

%%%%%%%%%%%%%%%%%%%%%%%%%%%%%%%%%%%%%%%%%%%%%%%%%%%%%%%%%%%%%%%%%%%%%%%%%%%%%%%%

%%%%%%%%%%%%%%%%%%%%%%%%%%%%%%%%%%%%%%%%%%%%%%%%%%%%%%%%%%%%%%%%%%%%%%%%%%%%%%%%

%%%%%%%%%%%%%%%%%%%%%%%%%%%%%%%%%%%%%%%%%%%%%%%%%%%%%%%%%%%%%%%%%%%%%%%%%%%%%%%%

%%%%%%%%%%%%%%%%%%%%%%%%%%%%%%%%%%%%%%%%%%%%%%%%%%%%%%%%%%%%%%%%%%%%%%%%%%%%%%%%

\bibliographystyle{IEEEtran}
\bibliography{reference}

% \newpage


% Latent conditioning formulation

% \begin{equation}
% \begin{aligned}
% p_{\theta}(z\mid\mathcal{C}_t)
% &=\mathcal{N}(\mu_p,\operatorname{diag}(\sigma_p^2)),\\
% q_{\theta}(z\mid\mathcal{C}_t,Y_t)
% &=\mathcal{N}(\mu_q,\operatorname{diag}(\sigma_q^2)).
% \end{aligned}
% \label{eq:latent_conditioning}
% \end{equation}

% Learnable gate setting, we use a signma function as the learnable gate, and it is trained during the policy learning stage.

% \begin{equation}
% g
% =
% \sigma(\gamma)
% \end{equation}

% Contrastive loss formulating

% \begin{equation}
% \ell_{v\rightarrow h}^{(i)}
% =
% -\frac{1}{B}
% \sum_{b=1}^{B}
% \log
% \frac{
% \exp
% \left(
% \operatorname{sim}
% (v_{b,i},h_{b,i})/\tau
% \right)
% }{
% \displaystyle
% \sum_{b'=1}^{B}
% \exp
% \left(
% \operatorname{sim}
% (v_{b,i},h_{b',i})/\tau
% \right)
% },
% \end{equation}
% where \(B\) denotes the batch size, \(\tau\) is the temperature coefficient, and \(\operatorname{sim}(\cdot,\cdot)\) denotes cosine similarity. The tactile-to-visual loss \(\ell_{h\rightarrow v}^{(i)}\) is defined analogously.

% KL Loss Design

% \begin{equation}
% \mathcal{L}_{\mathrm{KL}}
% =D_{\mathrm{KL}}\!\left[
% q_{\theta}(z\mid\mathcal{C}_t,Y_t)
% \,\middle\|\,
% p_{\theta}(z\mid\mathcal{C}_t)
% \right].
% \end{equation}

% Policy learning hyperparameter:

% \begin{equation}
% \lambda_{\mathrm{cross}}
% =
% \lambda_{\mathrm{temp}}
% =
% \lambda_{\mathrm{cons}}
% =
% 1,
% \qquad
% \tau
% =
% 0.07.
% \end{equation}

\end{document}